\documentclass[letterpaper, 10 pt, conference]{ieeeconf}
\IEEEoverridecommandlockouts
\usepackage{multirow}
\usepackage{multicol} 
\usepackage{makecell}
\usepackage{hyperref}
\usepackage{cite}
\usepackage{graphicx}
\usepackage{float}
\usepackage{amsfonts}
\usepackage{amsmath}
\usepackage{algorithm}
\usepackage{algpseudocode}
\usepackage{tabularx}
\usepackage{booktabs}
\usepackage{cleveref}
\usepackage[normalem]{ulem}
\usepackage{listings}
\usepackage{textcomp}
\usepackage{amssymb} 

\title{\LARGE \bf
LVMem: Image-Augmented Memory for Long-Horizon Question Answering
}

\author{%
    Victoria Khoruzhaya$^{1}$, 
    Maksim Kalmykov$^{1}$, 
    Tuong Do Minh$^{1}$, 
    Tatiana Zemskova$^{1,2}$, 
    Yixin Zhu$^{3}$, 
    Dmitry Yudin$^{1,2}$%
    \thanks{$^{1}$Moscow Independent Research Institute of Artificial Intelligence, Moscow, Russia}%
    \thanks{$^{2}$Cognitive AI Systems Lab, Moscow, Russia}%
    \thanks{$^{3}$Peking University, Beijing, China}%
}

\begin{document}

\maketitle

% ---------------------------------------------------------------

\begin{abstract}

Reliable language-guided perception is essential for embodied agents, where questions about objects, events, and their locations must be answered from long observation histories. Unlike conventional visual question answering, realistic robotic settings require reasoning over extended temporal horizons and persistent scene memory. We build upon the view-based memory paradigm and propose an efficient framework for long-horizon spatio-temporal question answering. While existing view-based systems typically store either text captions or visual embeddings, we extend the ReMEmbR-style memory with image paths that link each stored description to its original visual content. To make the stored annotations more informative, we replace uniform sampling during memory construction with an optical-flow-based frame selection strategy (PSFR) and store an object-centric descriptive scene representation instead of plain captions. We further equip a VLM-based agent with an adaptive retrieval tool that returns the image associated with the most relevant caption, allowing the agent to decide when visual inspection is needed. On the OC-NaVQA benchmark, LVMem outperforms ReMEmbR on binary and positional metrics with both backbones and matches or exceeds a 4D scene graph DAAAM baseline on binary accuracy, while operating on a lightweight view-based memory that requires no explicit graph construction. Our code is available at \href{https://github.com/comprock15/LVMem}{https://github.com/comprock15/LVMem}.

\end{abstract}

% ---------------------------------------------------------------

\section{Introduction}

Embodied agents that operate over hours or days accumulate vast streams of visual experience. A warehouse robot, for example, may be asked: ``Which shelf had the damaged box?'' or ``What did the loading dock look like when you passed it?''. Answering such queries goes beyond retrieving a text snippet: the agent must recall what it actually saw, where, and when. This demands a memory that preserves visual evidence across long horizons, remains spatially and temporally consistent, and can be queried efficiently as it grows.

\begin{figure}
    \centering
    \includegraphics[width=1.0\linewidth]{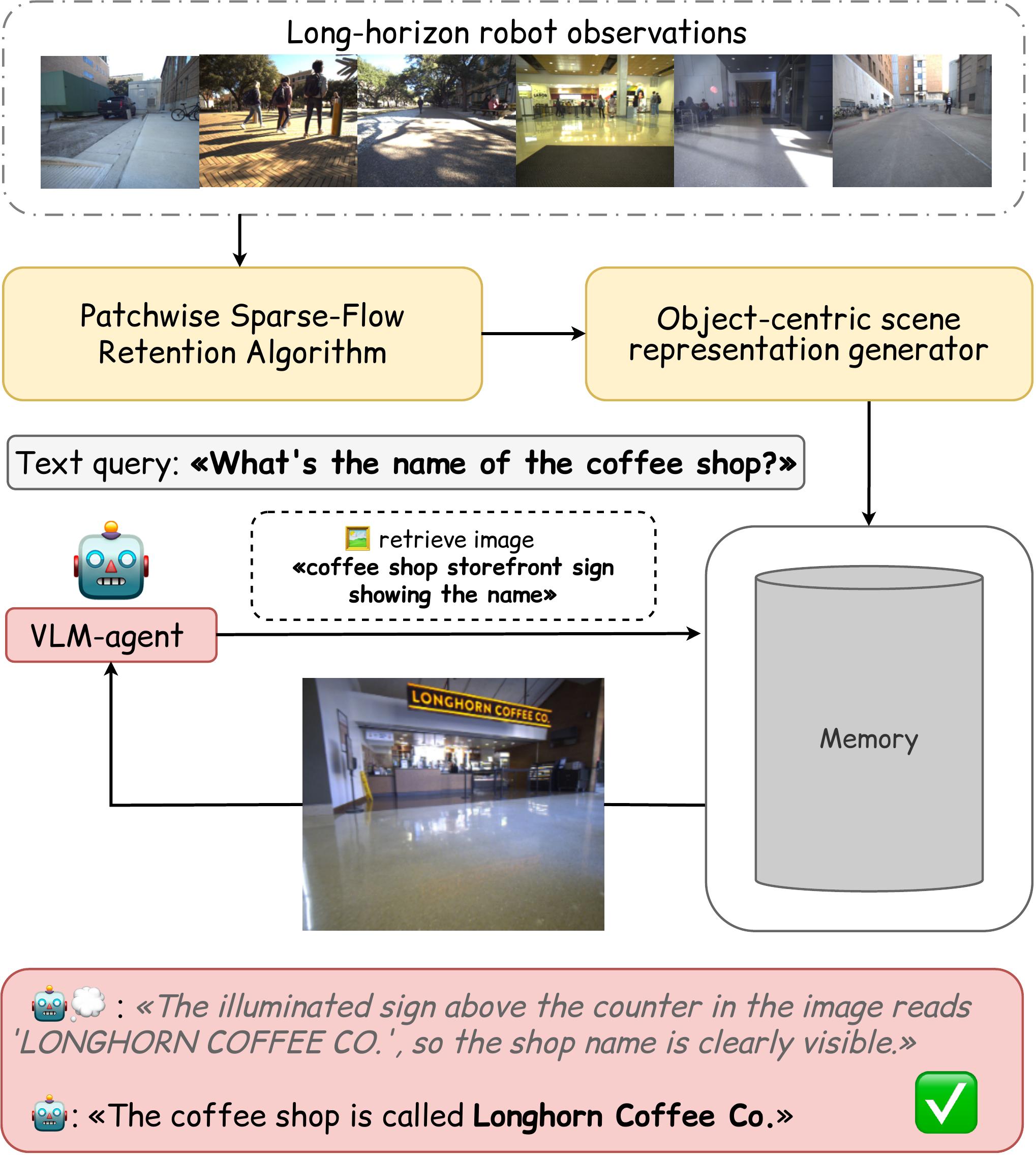}
    \caption{Proposed LVMem approach. A long video sequence is divided into fixed-length segments, from which PSFR selects the most informative frames. An object-centric textual scene representation generator produces one scene description per segment from its selected frames, which is stored in memory alongside the segment's timestamp, pose, and image path. For a given query, the VLM agent calls the image retrieval tool to obtain the image linked to the most relevant description, enabling joint reasoning over visual and textual evidence.}
    \label{fig:ga}
\end{figure}

We formalize the problem of long-horizon spatio-temporal question answering as follows. Given a continuous stream of RGB observations with associated poses and timestamps, and a natural language query about objects, events, or states in the environment, the goal is to produce an accurate answer. This requires building a memory representation from the observation stream and reasoning jointly over visual appearance, spatial locations, and temporal ordering.

This challenge becomes especially difficult in realistic robotic scenarios, where the environment evolves over time and the robot must reason over extended temporal horizons. A robot may observe the same object from multiple viewpoints, encounter changes in object locations, and need to combine information collected at different moments in time. Multimodal large language models have demonstrated strong capabilities in visual reasoning and language understanding~\cite{qi2025gpt4scene, wu2026st}, but they still struggle with long video sequences~\cite{wu2024longvideobench, hourvideo}, persistent scene memory~\cite{wang2025spatial4d}, and efficient retrieval of relevant information from large observation histories.

Structured scene representations offer a promising direction for addressing these limitations. In particular, scene graphs provide a compact and interpretable representation of objects, their attributes and interactions~\cite{jiao2022sequential}, and have been extended to dynamic 4D settings where nodes correspond to entities, edges encode semantic and spatial relations, and temporal information captures scene evolution~\cite{nguyen2026event, mdfaa2025vl}. Recent approaches such as DAAAM~\cite{gorlo2025daaam} have explored 4D scene graphs for robotics, yet they rely on explicit geometric reconstruction and per-object description generation. Such representations are attractive for robotics because they support persistent memory and enable reasoning across long horizons. However, constructing and maintaining explicit 4D scene graphs in large-scale environments remains computationally demanding.

An alternative line of work stores annotated observations in a structured memory index, such as a vector database, which can be queried efficiently without explicit 3D reconstruction. Recent approaches like ReMEmbR~\cite{anwar2025remembr} have shown promising results on spatio-temporal question answering using this paradigm, achieving efficient retrieval through text-based indexing of frame annotations. However, these methods rely solely on textual descriptions for retrieval and reasoning, which limits the agent's ability to verify visual details or resolve ambiguities that require direct inspection of the original image content. 

In this work, we build upon the view-based memory paradigm of ReMEmbR~\cite{anwar2025remembr} and propose an extension that improves long-horizon question answering by providing the agent with access to visual evidence beyond textual descriptions and by selecting more informative frames during memory construction. 

Our contribution can be summarized as follows:
\begin{itemize}
    \item We introduce LVMem, a view-based framework for long-horizon spatio-temporal question answering over recorded robot observations. LVMem represents observations as object-centric scene descriptions and links each to its source image, timestamp, and pose, equipping a VLM agent with a selectively invoked visual retrieval tool. This enables the agent to inspect visual details omitted from the stored descriptions while retaining text-, position-, and time-based access to memory, without explicit 3D reconstruction.
    \item We propose motion-guided memory construction under a fixed frame budget by integration of the existing Patchwise Sparse-Flow Retention (PSFR) method into segment-level scene description generation and evaluate it against uniform sampling with identical segment lengths and frame counts. With six frames per 10-second segment, PSFR improves binary accuracy by 8.96 percentage points and reduces mean positional error from 89.98 m to 68.16 m, demonstrating the importance of frame selection for downstream memory-based question answering.
    \item  Through comparisons with ReMEmbR and DAAAM methods and ablations on OC-NaVQA, we examine how frame selection, visual retrieval, text retrieval depth, and scene description configuration affect answer quality. LVMem improves binary and free-form text accuracy over ReMEmbR by 10.95 and 15.15 percentage points with Qwen3.5-9B, and by 19.90 and 5.05 points with GPT-5-Mini, respectively. Our ablations further show that visual augmentation depends on retrieval context: additional images can improve descriptive answers while degrading positional estimates or destabilizing temporal predictions.
\end{itemize}

% ---------------------------------------------------------------

\section{Related work}

Long-horizon embodied question answering requires reasoning over extended observation histories, where relevant evidence may be sparse and distributed across time. Existing approaches differ mainly in how they represent, retain and retrieve this information. 

\subsection{Long-Horizon Question Answering}

Recent MLLMs perform well on short clips but degrade as observation histories grow beyond the available context window. VideoARM~\cite{yin2025videoarm} performs agentic reasoning over a hierarchical memory for long-form video understanding. Benchmarks such as FindingDory~\cite{findingdory} and HourVideo~\cite{hourvideo} highlight the difficulty of retrieving sparse
evidence over long temporal horizons. Spatial reasoning remains challenging: GPT4Scene~\cite{qi2025gpt4scene} and ST-Think~\cite{wu2026st} study scene-level and 4D reasoning, while Spatial4D-Bench~\cite{wang2025spatial4d} and Video-MSR~\cite{zhu2026video} expose persistent weaknesses in spatio-temporal reasoning. Other approaches encode structure directly into video representations: Video-3D LLM~\cite{zheng2025video} introduces 3D-aware position encoding, and LITA~\cite{huang2024lita} uses explicit time tokens for temporal localization. Both require end-to-end training tied to a specific backbone, and such representations are typically consumed within the model context rather than retained as persistent, independently queryable memory. OnlineSI~\cite{liu2026onlinesi} moves toward persistent spatial memory through incremental semantic reconstruction, but its bounded representation is still passed directly to the language model rather than selectively queried.

\subsection{Scene Graphs and Knowledge Representations for Robotics}
\label{sec:scene-graphs}
Scene graphs explicitly organize observations into persistent spatial and semantic structure. HOV-SG~\cite{amuya2024hovsg} constructs hierarchical open-vocabulary graphs over floors, rooms, and objects, while ConceptGraphs~\cite{gu2024conceptgraphs} associates multi-view 3D objects with open-vocabulary semantic descriptors and inter-object relations. Clio~\cite{maggio2024clio} extends hierarchical scene graphs to online, task-driven construction, and Khronos~\cite{schmid2024khronos} models
short-term motion together with long-term environmental change. GraphEQA~\cite{saxena2025grapheqa} combines an online hierarchical scene graph with task-relevant images for embodied question answering. Among structured-memory approaches, DAAAM~\cite{gorlo2025daaam} builds a hierarchical 4D scene graph for long-horizon spatio-temporal QA, while EGG~\cite{nguyen2026event} introduces explicit event nodes and VL-KnG~\cite{mdfaa2025vl} constructs spatio-temporal knowledge graphs from video without explicit 3D reconstruction. Such graph-based representations provide rich global structure but require explicit graph construction or maintenance.

\subsection{Knowledge Retrieval and Agentic Reasoning over Structured Memory}

Persistent memory is only useful if relevant evidence can be retrieved efficiently. WorldMM~\cite{yeo2025worldmm}, EventMemAgent~\cite{wen2026eventmemagent}, and StreamReady~\cite{azad2026streamready} use episodic, semantic, or hierarchical memory for long-form reasoning, focusing on temporal abstraction rather than spatially grounded representations. ReMEmbR~\cite{anwar2025remembr} stores VLM-generated captions in a vector database and uses an iterative agent for text-, position-, and time-based retrieval, but the agent reasons only over stored textual descriptions. STaR~\cite{yuan2026star} additionally retrieves keyframes for visual inspection through task-conditioned selection. Graph-based memories such as DAAAM~\cite{gorlo2025daaam}, EGG~\cite{nguyen2026event}, and VL-KnG~\cite{mdfaa2025vl} combine structured representations with agentic or hybrid retrieval. Beyond retrieval, Surprise-Gated Memory~\cite{gorlo2026worth} studies selective episode retention, while UQ-DAAAM~\cite{zhang2026uq-daaam} quantify uncertainty over object descriptions and actively refine them. A tension persists: text-based memories scale efficiently but may discard visual detail, whereas multimodal memories retain richer evidence that must be selectively retrieved and integrated. LVMem retains the lightweight ReMEmbR representation while linking retrieved descriptions back to their source images, enabling a study when visual evidence actually helps.

% ---------------------------------------------------------------

\section{Method}

\begin{figure*}
    \centering
    \includegraphics[width=0.95\linewidth]{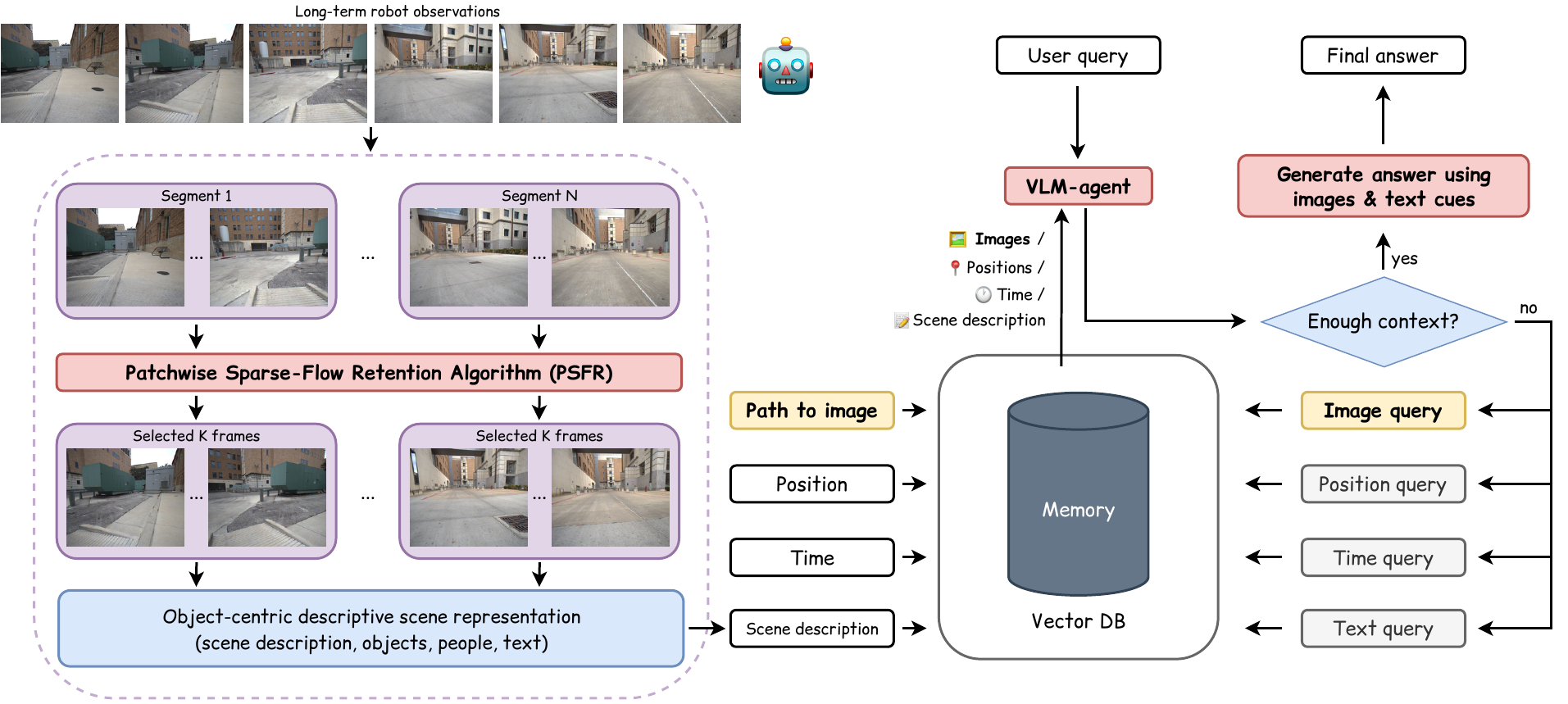}
    \caption{Overview of the proposed pipeline. Each temporal segment is processed by PSFR, which selects the most informative frames; the first selected frame is stored as an image path alongside the resulting object-centric scene description. At query time, the VLM agent invokes a visual retrieval tool that returns the image paired with the most relevant description, enabling joint reasoning over textual and visual evidence to reason jointly over both modalities.}
    \label{fig:proposed-pipeline}
\end{figure*}

We present a framework for long-horizon spatio-temporal question answering that extends the existing view-based memory system of ReMEmbR~\cite{anwar2025remembr}. Rather than redesigning the memory from scratch, we augment it in three ways. First, we replace uniform frame sampling during memory construction with an optical-flow-based selection strategy (PSFR)~\cite{zemskova2026focusgraph}. Second, we replace the plain captions with an object-centric descriptive scene representation and augment the memory entries with image paths that point to the corresponding source frames. Finally, we equip the VLM agent with an adaptive retrieval tool that returns the images corresponding to the most relevant descriptions, enabling joint reasoning over textual and visual evidence (see Fig.~\ref{fig:proposed-pipeline}).

\subsection{Memory Construction}

We follow the memory-building pipeline of ReMEmbR~\cite{anwar2025remembr}, which splits a video stream $S$ into fixed-length temporal segments $\{s_1, \dots, s_L\}$, each spanning $T$ seconds. We introduce two modifications to this pipeline.

First, instead of uniform frame sampling within each segment, we apply PSFR~\cite{zemskova2026focusgraph}, a training-free optical-flow-based method that selects the most informative frames while discarding static or redundant content. Let segment $s_i$ contain $m_i$ frames $F_i = \{f_{i,1}, \dots, f_{i,m_i}\}$. Uniform sampling selects $K$ frames at equal intervals, while PSFR selects a subset $\mathcal{F}_i \subset F_i$ of the same size $K$, using sparse optical flow to identify frames with strong structural change. Both methods therefore return exactly $K$ frames per segment, which allows a direct comparison under identical frame counts (see Table~\ref{tab:ablation_psfr}) and isolates the selection criterion from the frame budget.

Second, we introduce the \emph{object-centric descriptive scene representation}, a textual description of a segment designed for retrieval. A vision-language model $\phi_\theta$, instructed by a structured prompt $\pi$, maps them to four fields,
\begin{equation}
  d_i = \phi_\theta(\mathcal{F}_i;\pi) = \big(\sigma_i,\ \mathcal{O}_i,\ \mathcal{H}_i,\ \mathcal{X}_i\big),
  \label{eq:fields}
\end{equation}
where $\sigma_i$ is the scene type (with the weather, for outdoor scenes), $\mathcal{O}_i$ the visible objects, $\mathcal{H}_i$ the people with their activity and clothing, and $\mathcal{X}_i$ the readable text, quoted verbatim. Since a model asked to describe a scene tends to summarize it, omitting the entities a later query will ask about, the object field is specified as a set of attributed entities,
\begin{equation}
\begin{aligned}
  \mathcal{O}_i &= \big\{(w_j, c_j, n_j)\big\}_{j}, \qquad w_j \notin \mathcal{G},\\
  c_j &\in \mathcal{C} \cup \{\bot\}, \qquad n_j \in \mathbb{N},
\end{aligned}
\label{eq:objects}
\end{equation}
where $w_j$ is a noun, $c_j$ its color from a fixed palette $\mathcal{C}$, $n_j$ its count, $\bot$ the absence of a color and $\mathcal{G}$ a list of generic nouns (\textit{object}, \textit{item}, \textit{device}, etc.). The representation is thus \emph{enumerative} (distinct entities are listed separately, with repeated ones merged under a count), \emph{attributive} (color and count are bound to their noun), \emph{specific} ($w_j \notin \mathcal{G}$), and \emph{positive}: a field with no visible content is omitted rather than negated, since a statement such as ``no ATM is visible'' lies close in embedding space to a query about the same entity and would be retrieved while containing no answer.

The field labels are then stripped and empty fields dropped, yielding a prose description $\tau_i$. Generating with structure but storing without it is deliberate: requesting prose directly causes the model to summarize and drop entities, while retaining the labels would place identical tokens in every entry and dilute the embedding. Each segment is stored as the record
\begin{equation}
  r_i = \big(E(\tau_i),\ \tau_i,\ p_i,\ t_i,\ u_i\big),
  \label{eq:record}
\end{equation}
where $E$ is the mxbai-embed-large-v1 model~\cite{emb2024mxbai}, $p_i$ and $t_i$ are the pose and timestamp, and $u_i$ is the path to the first frame of $\mathcal{F}_i$.

We further augment the memory index with an image path column. Since PSFR discards static and redundant frames, retained frames contain non-trivial visual content, so we use the first one as a deterministic visual anchor without an additional ranking step. We store file paths rather than base64-encoded images to keep the index compact; the image data itself remains on disk, ensuring fast retrieval.

\subsection{Retrieval with Visual Augmentation}

We adopt the ReMEmbR retrieval architecture~\cite{anwar2025remembr}, which provides a tool-calling agent with access to text-based, position-based, and time-based retrieval over the vector database. All text-based search reduces to a top-$k$ query over the stored embeddings. Given a query $q$, retrieval returns the $k$ segments with the smallest $L_2$ distance to the query embedding,
\begin{equation}
  \mathcal{R}_k(q) = \operatorname*{arg\,top\text{-}k}_{i \in \{1, \dots, L\}}\ \big\|E(q) - e_i\big\|_2,
  \label{eq:topk}
\end{equation}
where $e_i = E(\tau_i)$ is the embedding of the $i$-th segment. The existing text tool then returns the corresponding descriptions,
\begin{equation}
  \mathrm{TextRetrieve}(q) = \{\tau_i\}_{i \in \mathcal{R}_k(q)}.
  \label{eq:textretrieve}
\end{equation}

We extend the toolset with a novel image retrieval tool, which accepts a natural language query and returns the image path corresponding to the most semantically relevant description. Internally, it performs the same top-$k$ search but returns image paths instead of text,
\begin{equation}
  \mathrm{ImageRetrieve}(q) = \{u_i\}_{i \in \mathcal{R}_M(q)},
  \label{eq:imageretrieve}
\end{equation}
where $M$ is a configurable parameter and $u_i$ is the anchor frame path from Eq.~\eqref{eq:record}.

As in the original ReMEmbR, the agent may invoke retrieval tools iteratively, up to a configurable number of steps. The agent can adaptively decide when to invoke the image retrieval tool: if textual descriptions are sufficient to answer a query, it proceeds without visual retrieval; if the query requires fine-grained visual details, such as reading signs or distinguishing similar objects, it calls the image retrieval tool to obtain the relevant images.

\subsection{Visual-Enhanced Answer Generation}

At the final reasoning step, we replace the LLM of ReMEmbR with a VLM that jointly processes retrieved texts and images. Before calling the VLM, the agent checks whether the image retrieval tool was invoked during the retrieval phase. If it was and the corresponding image paths were collected, these images are loaded and attached to the prompt; otherwise, generation proceeds in text-only mode.

Concretely, let $x$ denote the user's question and $\mathcal{Q} = \{q_1, \dots, q_m\}$ the set of retrieval sub-queries issued by the agent while answering $x$. Let
\[
  \mathcal{U} = \bigcup_{q \in \mathcal{Q}} \mathrm{ImageRetrieve}(q)
\]
be the set of unique image paths returned across all such calls. Each image is downscaled and re-encoded before being passed to the VLM: we rescale the longer side to at most $h_{\max} = 768$ pixels and save as JPEG with quality $\kappa = 80$, which reduces both the token cost and the payload size while preserving sufficient visual information. The resulting bytes are base64-encoded and attached to the prompt together with the textual context. We set the image detail level to \texttt{low} to further reduce token usage.

The VLM then produces a structured answer in JSON format with fields for text, binary, position, time, and duration, matching the output format of ReMEmbR. If the API returns a context-length error, the agent returns a default empty response instead of failing.

% ---------------------------------------------------------------

\section{Experiments}

\subsection{Dataset} Following prior work~\cite{anwar2025remembr, gorlo2025daaam}, we evaluate our approach on the OC-NaVQA benchmark~\cite{gorlo2025daaam}. The dataset comprises 7 video sequences from the CODa dataset~\cite{zhang2024coda}, featuring 210 questions about the robot's long-horizon navigation experience. Questions are categorized into spatial, temporal (point-in-time and duration), and descriptive (binary and text) types. For evaluation, we report Binary Accuracy (\%), Text Accuracy (\%, LLM-as-a-judge with Qwen3.5-0.8B), Positional Error (meters), and Temporal and Duration Errors (minutes) for all categories.

\subsection{Baselines} We compare our approach against two baselines: ReMEmbR~\cite{anwar2025remembr} in its original configuration (VILA1.5-3b~\cite{lin2024vila} captioner, original prompt) and DAAAM~\cite{gorlo2025daaam}. LVMem instead uses the object-centric descriptive scene representation (Qwen3.5-9B, structured prompt) 
and its retrieval tool returns the images linked to the 5 most
relevant scene representations. We evaluate all methods with two VLM backbones: Qwen3.5-9B and GPT-5-Mini, to analyze the impact of the reasoning agent. We use SGLang~\cite{zheng2024sglang} for local inference of open-source VLMs and OpenRouter for API-based access to proprietary models.

\subsection{Results}  

\begin{table*}[t]
\centering
\small
\caption{Comparison on OC-NaVQA. Mean $\pm$ std. dev. over 3 runs. ReMEmbR uses its original configuration, LVMem our proposed representation. Segment parameters are identical (3 seconds, 6 frames). Latency is measured on a single H100 GPU and reported only for local inference, as API-based runs incur variable network and queueing overhead.}
\begin{tabular}{l|ccccc|c}
\toprule
\textbf{Method} & \textbf{Binary [\%] $\uparrow$} & \textbf{Text [\%] $\uparrow$} & \textbf{Pos. [m] $\downarrow$} & \textbf{Temp. [min] $\downarrow$} & \textbf{Dur. [min] $\downarrow$} & \textbf{Latency [s] ↓} \\
\midrule
ReMEmbR (GPT-5-Mini) & 56.72$\pm$5.38 & 36.36$\pm$3.03 & 52.69$\pm$5.49 & 5.19$\pm$1.01 & \uline{3.66$\pm$0.99} & - \\
DAAAM (GPT-5-Mini) & 73.30$\pm$2.94 & \textbf{42.30$\pm$21.35} & 56.37$\pm$6.38 & \textbf{3.83$\pm$1.50} & \textbf{2.58$\pm$1.72} & -  \\
LVMem (Ours, GPT-5-Mini) & \textbf{76.62$\pm$3.76} & \uline{41.41$\pm$6.31} & \textbf{34.68$\pm$0.36} & 4.81$\pm$0.61 & 4.88$\pm$0.51 & - \\
\midrule
ReMEmbR (Qwen3.5-9B) & 65.17$\pm$3.11 & 24.24$\pm$3.03 & 68.67$\pm$12.76 & 16.95$\pm$20.46 & 4.96$\pm$2.22 & 19.19$\pm$5.05 \\
DAAAM (Qwen3.5-9B) & 43.50$\pm$10.97 & 23.09$\pm$2.97 & 65.92$\pm$7.92 & 7.86$\pm$1.98 & 3.87$\pm$2.99 & 30.67$\pm$38.76 \\
LVMem (Ours, Qwen3.5-9B) & \uline{76.12$\pm$2.59} & 39.39$\pm$3.03 & \uline{50.12$\pm$4.96} & \uline{4.23$\pm$0.54}  & 3.67$\pm$2.30 & 20.91$\pm$5.33 \\
\bottomrule
\end{tabular}
\label{tab:main_results}
\end{table*}

Table~\ref{tab:main_results} reports the comparison on OC-NaVQA. LVMem outperforms ReMEmbR on binary and text accuracy with both backbones, and reduces positional error substantially in both configurations. With GPT-5-Mini, LVMem attains the highest binary accuracy among all methods and remains competitive on text accuracy, though its duration error is higher than DAAAM's, while its temporal error falls between ReMEmbR and DAAAM. With Qwen3.5-9B, LVMem improves over ReMEmbR on binary, text, positional, and temporal metrics; the two methods are close on duration
errors.

DAAAM with GPT-5-Mini achieves the strongest text accuracy and the lowest temporal and duration errors, but it is highly sensitive to the choice of backbone: switching to Qwen3.5-9B substantially degrades its binary and text accuracy, below both ReMEmbR and LVMem. LVMem, in contrast, remains stable across backbones and improves with Qwen3.5-9B
on binary and positional metrics, while ReMEmbR shows the opposite
trend, with GPT-5-Mini outperforming Qwen on text. This suggests that
the optimal backbone depends on both the memory representation and the
question type. Overall, LVMem offers the most consistent performance across backbones, while DAAAM achieves the strongest results only with GPT-5-Mini.

\begin{figure}
    \centering
    \includegraphics[width=1.0\linewidth]{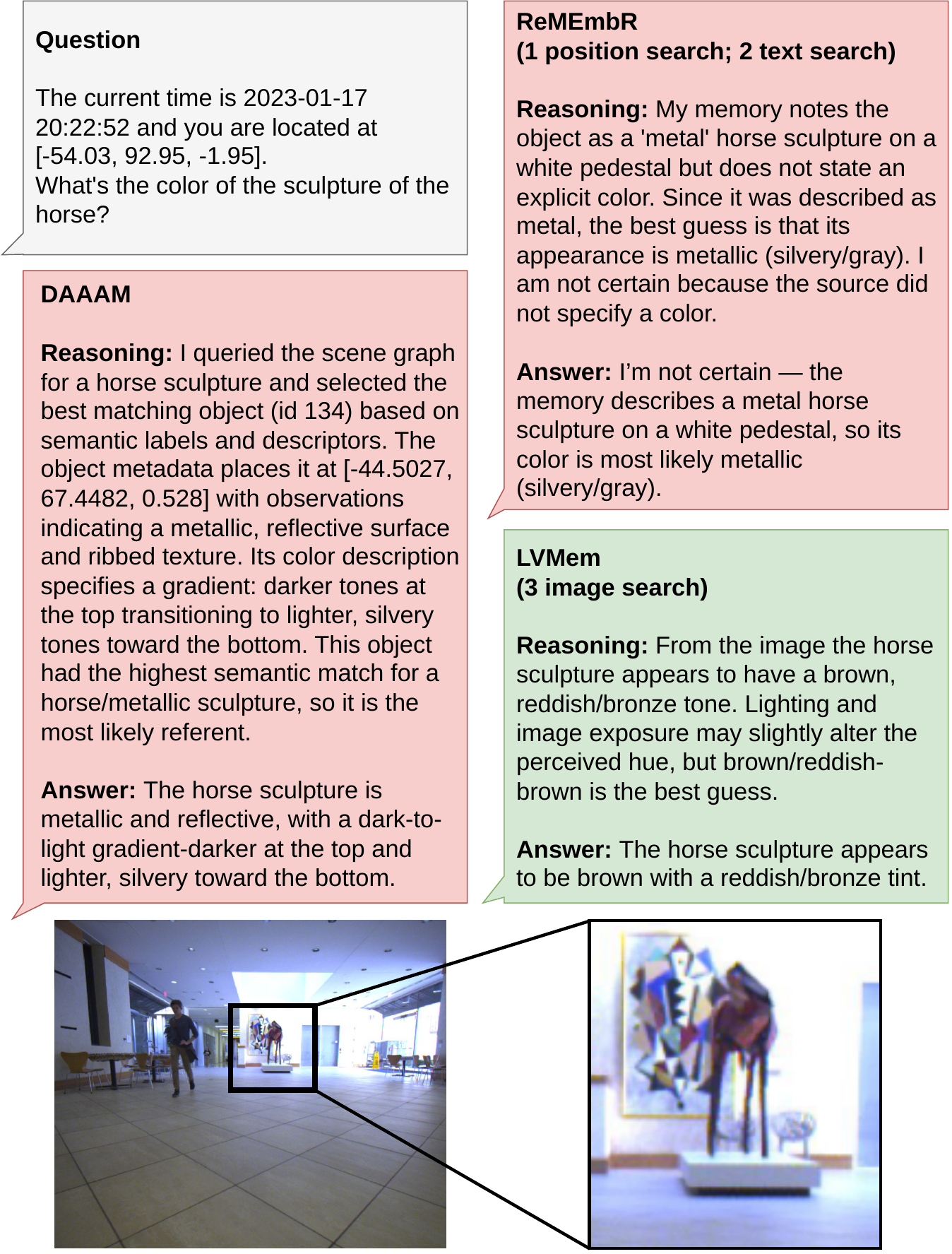}
    \caption{A question asking for the color of a horse sculpture. Text-based baselines describe the sculpture as ``metal'' and ``reflective'' and cannot determine its color. LVMem retrieves the corresponding frame and answers correctly based on the visible brown tone.}
    \label{fig:comparison}
\end{figure}

Figure~\ref{fig:comparison} illustrates a failure mode of text-only memory: when a question targets a fine-grained visual attribute, the stored description may omit it even if the corresponding observation was recorded. Here, both baselines store a description of the sculpture but neither mentions its color, and DAAAM's scene graph node, though linked to the correct object, describes only its reflective surface. LVMem avoids this failure because the adaptive retrieval tool lets the agent fetch the source frame and answer from the image itself, rather than from the stored text alone.

\subsection{Ablation studies} 

\subsubsection{PSFR vs uniform sampling}

\begin{table*}[t]
\centering
\small
\caption{Effect of PSFR on OC-NaVQA performance for LVMem with Qwen3.5-9B. Segment parameters: 10 seconds, 6 frames}
\begin{tabular}{c|ccccc}
\toprule
\textbf{Sampling} & \textbf{Binary [\%] $\uparrow$} & \textbf{Text [\%] $\uparrow$} & \textbf{Pos. [m] $\downarrow$} & \textbf{Temp. [min] $\downarrow$} & \textbf{Dur. [min] $\downarrow$} \\
\midrule 
uniform & 55.72$\pm$5.65 & 22.22$\pm$1.75 & 89.98$\pm$9.05 & 5.60$\pm$2.26 & 4.27$\pm$0.12 \\
PSFR & \textbf{64.68$\pm$4.80} & \textbf{25.25$\pm$1.75} & \textbf{68.16$\pm$12.96} & \textbf{4.71$\pm$2.66} & \textbf{3.83$\pm$0.99} \\
\bottomrule
\end{tabular}
\label{tab:ablation_psfr}
\end{table*}

Table~\ref{tab:ablation_psfr} isolates the effect of frame selection during memory construction. Both variants use the same segment parameters (10 seconds, 6 frames), the same visual retrieval tool returning one image, and the original ReMEmbR captioning configuration (VILA1.5-3b, original prompt). 

Replacing uniform sampling with PSFR consistently improves all metrics, with the largest gain on binary accuracy (8.96\%) and a smaller improvement in text accuracy. Positional error drops substantially, and temporal and duration errors also decrease. The improvements are consistent across all three runs, as indicated by the standard deviations.

These results confirm that the quality of stored descriptions depends directly on which frames are selected. By prioritizing frames with significant visual content and discarding static or redundant ones, PSFR produces more informative descriptions, which in turn improve retrieval and downstream question answering.

\subsubsection{Retrieval depth and visual augmentation}

\begin{table*}[t]
\centering
\small
\caption{Joint effect of text retrieval depth $k$ and the number of images $M$ on OC-NaVQA with LVMem (Qwen3.5-9B). In the \emph{top-M} mode, images from the top-$M$ retrieved descriptions are always appended directly to the generation prompt. In the \emph{tool} mode, the agent instead invokes the adaptive visual retrieval tool, which returns the images linked to the top retrieved descriptions. Segment parameters: 3 seconds, 6 frames.}
\begin{tabular}{ccc|ccccc}
\toprule
\textbf{Mode} & \textbf{$k$} & \textbf{Img} & \textbf{Bin. [\%] $\uparrow$} & \textbf{Text [\%] $\uparrow$} & \textbf{Pos. [m] $\downarrow$} & \textbf{Temp. [min] $\downarrow$} & \textbf{Dur. [min] $\downarrow$} \\
\midrule
top-M & 5  & 0 & 66.70$\pm$0.90 & 34.30$\pm$7.60 & \uline{60.30$\pm$0.20} & 4.16$\pm$0.23 & 5.92$\pm$2.02 \\
top-M  & 20 & 0 & 67.20$\pm$3.00 & 30.30$\pm$3.00 & \textbf{54.50$\pm$16.90} & \uline{3.39$\pm$0.67} & 4.53$\pm$0.12 \\
\midrule

tool & 5 & 1 & 69.65$\pm$1.72 & 28.28$\pm$1.75 & 68.15$\pm$4.72 & 5.43$\pm$1.69 & 4.99$\pm$1.73 \\
top-M  & 5  & 1 & 69.20$\pm$3.80 & \textbf{38.40$\pm$1.70} & 78.20$\pm$3.90 & \textbf{2.03$\pm$0.18} & \uline{3.79$\pm$0.10} \\
top-M  & 20 & 1 & 68.20$\pm$2.30 & 32.30$\pm$6.30 & 72.50$\pm$6.70 & 6.50$\pm$2.68 & 4.19$\pm$0.67 \\
\midrule

tool & 5 & 3 & 74.13$\pm$2.28 & 22.22$\pm$1.75 & 67.53$\pm$4.82 & 4.75$\pm$1.27 & 3.78$\pm$0.34 \\
top-M  & 5  & 3 & 73.00$\pm$1.70 & 33.30$\pm$3.00 & 77.10$\pm$8.00 & 4.43$\pm$0.77 & 4.34$\pm$0.67 \\
top-M  & 20 & 3 & 65.70$\pm$3.90 & \uline{37.70$\pm$4.20} & 62.60$\pm$3.10 & 5.66$\pm$0.17 & 4.73$\pm$0.38 \\
\midrule

tool & 5 & 5 & \uline{75.62$\pm$3.11} & 34.34$\pm$1.75 & 68.76$\pm$2.89 & 24.51$\pm$34.86 & \textbf{3.15$\pm$0.44} \\
top-M  & 5  & 5 & \textbf{77.30$\pm$2.50} & 34.10$\pm$1.20 & 71.70$\pm$3.70 & 4.73$\pm$0.46 & 4.25$\pm$1.03 \\
top-M  & 20 & 5 & 66.70$\pm$3.40 & 35.80$\pm$7.10 & 66.00$\pm$10.60 & 5.30$\pm$1.26 & 4.63$\pm$0.02 \\
\bottomrule
\end{tabular}
\label{tab:ablation_k_images}
\end{table*}

Table~\ref{tab:ablation_k_images} varies the number of documents returned by text retrieval ($k$) jointly with the number of images $M$, under two modes of supplying visual evidence. In the tool mode, the agent invokes the adaptive visual retrieval tool, which returns the images linked to the top retrieved descriptions. In the top-M mode, the tool is bypassed and images from the top-M retrieved descriptions are appended directly to the generation prompt. The $M=0$ rows use no images and thus serve as text-only baselines under identical memory construction. Descriptions, frame selection, and the agent are held fixed throughout.

Attaching images at shallow retrieval depth improves binary accuracy monotonically, with the gain growing as more images are added. At deep retrieval depth, however, the effect disappears: binary accuracy is flat across image counts. The two information sources therefore act as substitutes rather than complements — images supply information the agent otherwise lacks only when the textual context is shallow, and contribute little once the prompt already contains many descriptions.

Positional error follows the opposite pattern. It is lowest with no images, at both retrieval depths, and increases once images are attached at every non-zero setting. The effect is systematic rather than run-to-run variation. We hypothesize that when images are present, the VLM shifts toward visual grounding and produces less accurate spatial estimates than when reasoning from the positional metadata embedded in the retrieved descriptions.

Retrieval depth does not improve accuracy: moving from $k=5$ to $k=20$ leaves binary accuracy unchanged at every image count, and positional error improves only modestly. We therefore adopt $k=5$ in all remaining experiments.

Finally, we compare the adaptive tool against a variant that always attaches images from the last retrieved documents on every query. This non-adaptive variant attains competitive binary accuracy but suffers from a higher duration error. This confirms that selectively invoking the visual tool matters: indiscriminately attaching images to every query degrades temporal reasoning. Overall, a moderate number of images per tool call provides the best trade-off between accuracy and stability, while more images yield the highest binary accuracy at the cost of increased variance.

\subsubsection{Effect of the scene description models}

\begin{table*}[ht]
\centering
\small
\caption{Effect of different scene description models and prompts within our LVMem pipeline on OC-NaVQA at $k{=}5$. Images from top-$M$ retrieved documents are attached at generation; the visual retrieval tool is not used. Segment parameters: 3 seconds, 6 frames.}
\begin{tabular}{l|c|c|ccccc}
\toprule

\textbf{\makecell[l]{Scene description \\ model}} & \textbf{Prompt} & \textbf{Images} & \textbf{Binary [\%] $\uparrow$} & \textbf{Text [\%] $\uparrow$} & \textbf{Pos. [m] $\downarrow$} & \textbf{Temp. [min] $\downarrow$} & \textbf{Dur. [min] $\downarrow$} \\
\midrule
\multirow{4}{*}{VILA1.5-3b} & \multirow{4}{*}{ReMEmbR} & 0 & 66.70$\pm$0.90 & 34.30$\pm$7.60 & \textbf{60.30$\pm$0.20} & 4.16$\pm$0.23 & 5.92$\pm$2.02 \\
&  & 1 & 69.20$\pm$3.80 & \textbf{38.40$\pm$1.70} & 78.20$\pm$3.90 & \textbf{2.03$\pm$0.18} & \textbf{3.79$\pm$0.10} \\
&  & 3 & 73.00$\pm$1.70 & 33.30$\pm$3.00 & 77.10$\pm$8.00   & 4.43$\pm$0.77 & 4.34$\pm$0.67 \\
&  & 5 & \textbf{77.30$\pm$2.50} & 34.10$\pm$1.20 & 71.70$\pm$3.70 & 4.73$\pm$0.46 & 4.25$\pm$1.03 \\
\midrule
\multirow{4}{*}{Qwen3.5-9B} & \multirow{4}{*}{LVMem} & 0 & 76.10$\pm$2.10 & 31.10$\pm$1.50 & \textbf{43.60$\pm$3.00} & 15.98$\pm$22.23 & 4.46$\pm$1.43 \\
& & 1 & \textbf{81.00$\pm$5.40} & 31.80$\pm$3.90 & 53.30$\pm$4.60 & 3.82$\pm$1.10 & 4.68$\pm$0.65 \\
& & 3 & 76.90$\pm$0.90 & \textbf{34.80$\pm$1.70} & 60.00$\pm$3.80 & 5.62$\pm$2.78 & \textbf{4.20$\pm$1.11} \\
& & 5 & 78.00$\pm$1.90 & 30.30$\pm$5.50 & 54.60$\pm$5.40 & \textbf{3.56$\pm$0.70} & 4.76$\pm$0.94 \\
\bottomrule
\addlinespace[1ex]
\end{tabular}
\label{tab:ablation_captioner}
\end{table*}

Table~\ref{tab:ablation_captioner} examines the effect of the scene
description model by replacing VILA1.5-3b with Qwen3.5-9B while keeping all other components fixed ($k{=}5$, same embedding model, same agent). The two configurations differ both in model capacity and in prompt design: VILA receives a generic free-form prompt from the original ReMEmbR codebase, while Qwen is guided by a structured prompt that enforces explicit object enumeration with colors from a fixed palette, prohibits negations and hedging language, and limits the output to 110 words. The structured output is post-processed into continuous prose to ensure consistent embedding.

Binary accuracy improves with Qwen across all image settings, with the largest gain at one image. Positional error is also lower with Qwen at every image count, and both improvements exceed the run-to-run standard deviation, indicating a robust effect rather than an artifact of a particular configuration.

The gain can be traced to the coverage of stored entities: VILA descriptions omit objects such as ATMs and police emergency poles that Qwen descriptions include. For questions about such objects, VILA-based retrieval returns irrelevant documents, while Qwen-based retrieval finds the correct entry, cutting positional error by up to an order of magnitude.

Text accuracy, however, does not improve: VILA is higher in three of four image settings. Such questions require distinguishing between similar objects (e.g., ``the bike rack closest to the trash can''), and mentions of common objects such as bike racks increase by an order of magnitude in Qwen descriptions, diluting retrieval and returning a plausible but incorrect entry. The structured prompt thus improves the inclusion of rare objects at the cost of disambiguation among frequent ones.

Temporal error for Qwen at zero images is inflated by a single outlier of in one of the three runs; the remaining temporal and duration values are comparable between models and show no consistent advantage for either model.

% ---------------------------------------------------------------

\section{Limitations}

Our approach has several limitations. First, increasing the number of images returned by the visual retrieval tool does not monotonically improve performance. While additional images can help resolve queries that require fine-grained visual details, they also introduce noise: at five images per tool call, temporal error is dominated by outliers, and a non-adaptive variant that always attaches images to every query degrades duration estimation by several orders of magnitude. This confirms that selective invocation matters, but also that the number of images per call requires careful tuning.

Second, the observed gains are concentrated on binary and free-form text accuracy, where LVMem improves consistently over ReMEmbR. Duration estimation, in contrast, shows no consistent improvement across backbones: LVMem is better with Qwen3.5-9B but worse with GPT-5-Mini. This is expected, since our method targets visual details that are lost in the stored descriptions, and such details matter most for descriptive queries.

Finally, our method operates on a lightweight view-based memory and does not build an explicit 3D representation. This design choice reduces memory construction cost but may limit performance on questions that benefit from explicit spatial or temporal structure: LVMem improves over DAAAM on binary and positional metrics with some backbones, but remains behind on text accuracy and duration estimation.

% ---------------------------------------------------------------

\section{Conclusion}

We presented LVMem, a framework for long-horizon spatio-temporal question answering that extends the view-based memory of ReMEmbR with visual content and improved frame selection. We make three contributions: we augment the memory index with image paths that link each stored object-centric descriptive scene representation to its source frame, we introduce an adaptive visual retrieval tool that allows the agent to request images when textual descriptions are insufficient, and we replace uniform frame sampling with existing PSFR~\cite{zemskova2026focusgraph}, an optical-flow-based strategy that selects more informative frames for scene description generation.

On the OC-NaVQA benchmark, our method improves over ReMEmbR on binary, text, and positional metrics with both backbones, and substantially reduces temporal error, while remaining competitive with the 4D scene graph method DAAAM without requiring explicit graph construction. Ablations confirm that both PSFR and adaptive image retrieval contribute to these gains, and that indiscriminate image attachment degrades performance.

These results suggest that visual augmentation of view-based memory is a practical and effective direction for long-horizon question answering, improving scene perception for embodied agents.

\section*{Acknowledgment}

The authors used AI-based tools during the preparation of this manuscript. For text drafting, editing, and translation, the authors used DeepSeek (accessed via the web interface) and OpenAI models. For code generation,
debugging, and refactoring, the authors used DeepSeek, OpenAI Codex, and Claude. Table formatting and layout were also assisted by these tools. All AI-generated content was reviewed, verified, and edited by the authors, who take full responsibility for the content of this work.

{
    \small
    \bibliographystyle{IEEEtran}
    \bibliography{ref}

@article{qi2025gpt4scene,
  title={Gpt4scene: Understand 3d scenes from videos with vision-language models},
  author={Qi, Zhangyang and Zhang, Zhixiong and Fang, Ye and Wang, Jiaqi and Zhao, Hengshuang},
  journal={arXiv preprint arXiv:2501.01428},
  year={2025}
}

@inproceedings{wu2026st,
  title={St-think: How multimodal large language models reason about 4d worlds from ego-centric videos},
  author={Wu, Peiran and Liu, Yunze and Liu, Miao and Shen, Junxiao},
  booktitle={Proceedings of the IEEE/CVF Winter Conference on Applications of Computer Vision},
  pages={5174--5183},
  year={2026}
}

@article{wu2024longvideobench,
  title={Longvideobench: A benchmark for long-context interleaved video-language understanding},
  author={Wu, Haoning and Li, Dongxu and Chen, Bei and Li, Junnan},
  journal={Advances in Neural Information Processing Systems},
  volume={37},
  pages={28828--28857},
  year={2024}
}

@article{hourvideo,
  title={Hourvideo: 1-hour video-language understanding},
  author={Chandrasegaran, Keshigeyan and Gupta, Agrim and Hadzic, Lea M and Kota, Taran and He, Jimming and Eyzaguirre, Crist{\'o}bal and Durante, Zane and Li, Manling and Wu, Jiajun and Fei-Fei, Li},
  journal={Advances in Neural Information Processing Systems},
  volume={37},
  pages={53168--53197},
  year={2024}
}

@article{wang2025spatial4d,
  title={Spatial4D-Bench: A Versatile 4D Spatial Intelligence Benchmark},
  author={Wang, Pan and Liu, Yang and Wu, Guile and Corral-Soto, Eduardo R and Huang, Chengjie and Xu, Binbin and Bai, Dongfeng and Yan, Xu and Ren, Yuan and Chen, Xingxin and others},
  journal={arXiv preprint arXiv:2601.00092},
  year={2025}
}

@inproceedings{jiao2022sequential,
  title={Sequential manipulation planning on scene graph},
  author={Jiao, Ziyuan and Niu, Yida and Zhang, Zeyu and Zhu, Song-Chun and Zhu, Yixin and Liu, Hangxin},
  booktitle={2022 IEEE/RSJ International Conference on Intelligent Robots and Systems (IROS)},
  pages={8203--8210},
  year={2022},
  organization={IEEE}
}

@article{nguyen2026event,
  title={Event-Grounding Graph: Unified Spatio-Temporal Scene Graph From Robotic Observations},
  author={Nguyen, Phuoc and Verdoja, Francesco and Kyrki, Ville},
  journal={IEEE Robotics and Automation Letters},
  year={2026},
  publisher={IEEE}
}

@article{mdfaa2025vl,
  title={VL-KnG: Visual Scene Understanding for Navigation Goal Identification using Spatiotemporal Knowledge Graphs},
  author={Mdfaa, Mohamad Al and Lukina, Svetlana and Akhtyamov, Timur and Nigmatzyanov, Arthur and Nalberskii, Dmitrii and Zagoruyko, Sergey and Ferrer, Gonzalo},
  journal={arXiv preprint arXiv:2510.01483},
  year={2025}
}

@article{gorlo2025daaam,
  title        = {Describe Anything Anywhere At Any Moment},
  author       = {Gorlo, Nicolas and Schmid, Lukas and Carlone, Luca},
  journal      = {arXiv preprint arXiv:2512.00565},
  year         = {2025},
  eprint       = {2512.00565},
  archivePrefix= {arXiv},
  primaryClass = {cs.CV},
  url          = {https://arxiv.org/abs/2512.00565}
}

@inproceedings{anwar2025remembr,
  title={Remembr: Building and reasoning over long-horizon spatio-temporal memory for robot navigation},
  author={Anwar, Abrar and Welsh, John and Biswas, Joydeep and Pouya, Soha and Chang, Yan},
  booktitle={2025 IEEE International Conference on Robotics and Automation (ICRA)},
  pages={2838--2845},
  year={2025},
  organization={IEEE}
}

@article{yin2025videoarm,
  title={VideoARM: Agentic Reasoning over Hierarchical Memory for Long-Form Video Understanding},
  author={Yin, Yufei and Meng, Qianke and Chen, Minghao and Ding, Jiajun and Shao, Zhenwei and Yu, Zhou},
  journal={arXiv preprint arXiv:2512.12360},
  year={2025}
}

@article{findingdory,
  title={Findingdory: A benchmark to evaluate memory in embodied agents},
  author={Yadav, Karmesh and Ali, Yusuf and Gupta, Gunshi and Gal, Yarin and Kira, Zsolt},
  journal={arXiv preprint arXiv:2506.15635},
  year={2025}
}

@article{zhu2026video,
  title={Video-MSR: Benchmarking Multi-hop Spatial Reasoning Capabilities of MLLMs},
  author={Zhu, Rui and Shen, Xin and Wu, Shuchen and Miao, Chenxi and Yu, Xin and Li, Yang and Li, Weikang and Xia, Deguo and Huang, Jizhou},
  journal={arXiv preprint arXiv:2601.09430},
  year={2026}
}

@inproceedings{zheng2025video,
  title={Video-3d llm: Learning position-aware video representation for 3d scene understanding},
  author={Zheng, Duo and Huang, Shijia and Wang, Liwei},
  booktitle={Proceedings of the IEEE/CVF Conference on Computer Vision and Pattern Recognition},
  pages={8995--9006},
  year={2025}
}

@inproceedings{huang2024lita,
  title={Lita: Language instructed temporal-localization assistant},
  author={Huang, De-An and Liao, Shijia and Radhakrishnan, Subhashree and Yin, Hongxu and Molchanov, Pavlo and Yu, Zhiding and Kautz, Jan},
  booktitle={European Conference on Computer Vision},
  pages={202--218},
  year={2024},
  organization={Springer}
}

@article{liu2026onlinesi,
  title={OnlineSI: Taming Large Language Model for Online 3D Understanding and Grounding},
  author={Liu, Zixian and Chen, Zhaoxi and Pan, Liang and Liu, Ziwei},
  journal={arXiv preprint arXiv:2601.16538},
  year={2026}
}

@inproceedings{amuya2024hovsg,
  title        = {Hierarchical Open-Vocabulary 3D Scene Graphs for Language-Grounded Robot Navigation},
  author       = {Werby, Abdelrhman and Huang, Chenguang and B{\"u}chner, Martin and Valada, Abhinav and Burgard, Wolfram},
  booktitle    = {Robotics: Science and Systems (RSS)},
  year         = {2024},
  url          = {https://arxiv.org/abs/2403.17846}
}

@inproceedings{gu2024conceptgraphs,
  title={Conceptgraphs: Open-vocabulary 3d scene graphs for perception and planning},
  author={Gu, Qiao and Kuwajerwala, Ali and Morin, Sacha and Jatavallabhula, Krishna Murthy and Sen, Bipasha and Agarwal, Aditya and Rivera, Corban and Paul, William and Ellis, Kirsty and Chellappa, Rama and others},
  booktitle={2024 IEEE International Conference on Robotics and Automation (ICRA)},
  pages={5021--5028},
  year={2024},
  organization={IEEE}
}

@ARTICLE{maggio2024clio,
  author={Maggio, Dominic and Chang, Yun and Hughes, Nathan and Trang, Matthew and Griffith, Dan and Dougherty, Carlyn and Cristofalo, Eric and Schmid, Lukas and Carlone, Luca},
  journal={IEEE Robotics and Automation Letters}, 
  title={Clio: Real-Time Task-Driven Open-Set 3D Scene Graphs}, 
  year={2024},
  volume={9},
  number={10},
  pages={8921-8928},
  doi={10.1109/LRA.2024.3451395}}

@inproceedings{schmid2024khronos,
  title     = {Khronos: A Unified Approach for Spatio-Temporal Metric-Semantic SLAM in Dynamic Environments},
  author    = {Schmid, Lukas and Abate, Marcus and Chang, Yun and Carlone, Luca},
  booktitle = {Robotics: Science and Systems (RSS)},
  year      = {2024}
}

@inproceedings{saxena2025grapheqa,
  title     = {{GraphEQA}: Using {3D} Semantic Scene Graphs for Real-time Embodied Question Answering},
  author    = {Saxena, Saumya and Buchanan, Blake and Paxton, Chris and Liu, Peiqi and Chen, Bingqing and Vaskevicius, Narunas and Palmieri, Luigi and Francis, Jonathan and Kroemer, Oliver},
  booktitle = {Proceedings of the 9th Conference on Robot Learning},
  series    = {Proceedings of Machine Learning Research},
  volume    = {305},
  pages     = {2714--2742},
  year      = {2025},
  publisher = {PMLR}
}

@article{yeo2025worldmm,
  title={Worldmm: Dynamic multimodal memory agent for long video reasoning},
  author={Yeo, Woongyeong and Kim, Kangsan and Yoon, Jaehong and Hwang, Sung Ju},
  journal={arXiv preprint arXiv:2512.02425},
  year={2025}
}

@article{wen2026eventmemagent,
  title={EventMemAgent: Hierarchical Event-Centric Memory for Online Video Understanding with Adaptive Tool Use},
  author={Wen, Siwei and Wang, Zhangcheng and Zhang, Xingjian and Huang, Lei and Wu, Wenjun},
  journal={arXiv preprint arXiv:2602.15329},
  year={2026}
}

@article{azad2026streamready,
  title={StreamReady: Learning What to Answer and When in Long Streaming Videos},
  author={Azad, Shehreen and Vineet, Vibhav and Rawat, Yogesh Singh},
  journal={arXiv preprint arXiv:2603.08620},
  year={2026}
}

@article{yuan2026star,
  title={STaR: Scalable Task-Conditioned Retrieval for Long-Horizon Multi-Modal Robot Memory},
  author={Yuan, Mingfeng and Zhang, Hao and Mohammadi, Mahan and Li, Runhao and Shan, Jinjun and Waslander, Steven L},
  journal={IEEE Robotics and Automation Letters},
  volume={11},
  number={5},
  pages={5994--6001},
  year={2026},
  publisher={IEEE}
}

@article{gorlo2026worth,
  title={Worth Remembering: Surprise-Gated Robot Episodic Memory},
  author={Gorlo, Nicolas and Wise, Derek K and Speranzon, Alberto and Carlone, Luca},
  journal={arXiv preprint arXiv:2606.03787},
  year={2026}
}

@article{zemskova2026focusgraph,
  title={FocusGraph: Graph-Structured Frame Selection for Embodied Long Video Question Answering},
  author={Zemskova, Tatiana and Andryushenko, Solomon and Obrubov, Ilya and Khoruzhaia, Viktoriia and Eroshenko, Ekaterina and Derevyanka, Ekaterina and Yudin, Dmitry},
  journal={arXiv preprint arXiv:2603.04349},
  year={2026}
}

@online{emb2024mxbai,
  title={Open Source Strikes Bread - New Fluffy Embeddings Model},
  author={Sean Lee and Aamir Shakir and Darius Koenig and Julius Lipp},
  year={2024},
  url={https://www.mixedbread.ai/blog/mxbai-embed-large-v1},
}

@article{zhang2024coda,
  title={Toward robust robot 3-d perception in urban environments: The ut campus object dataset},
  author={Zhang, Arthur and Eranki, Chaitanya and Zhang, Christina and Park, Ji-Hwan and Hong, Raymond and Kalyani, Pranav and Kalyanaraman, Lochana and Gamare, Arsh and Bagad, Arnav and Esteva, Maria and others},
  journal={IEEE Transactions on Robotics},
  volume={40},
  pages={3322--3340},
  year={2024},
  publisher={IEEE}
}

@article{zhang2026uq-daaam,
  title={Remember with Confidence: Uncertainty Quantification for Spatio-temporal Memory with Probabilistic Guarantees},
  author={Zhang, Harry and Gorlo, Nicolas and Carlone, Luca},
  journal={arXiv preprint arXiv:2606.08277},
  year={2026}
}

@inproceedings{lin2024vila,
  title={Vila: On pre-training for visual language models},
  author={Lin, Ji and Yin, Hongxu and Ping, Wei and Molchanov, Pavlo and Shoeybi, Mohammad and Han, Song},
  booktitle={2024 IEEE/CVF Conference on Computer Vision and Pattern Recognition (CVPR)},
  pages={26679--26689},
  year={2024},
  organization={IEEE}
}

@article{zheng2024sglang,
  title={Sglang: Efficient execution of structured language model programs},
  author={Zheng, Lianmin and Yin, Liangsheng and Xie, Zhiqiang and Sun, Chuyue and Huang, Jeff and Yu, Cody H and Cao, Shiyi and Kozyrakis, Christos and Stoica, Ion and Gonzalez, Joseph E and others},
  journal={Advances in neural information processing systems},
  volume={37},
  pages={62557--62583},
  year={2024}
}
}

\end{document}